\documentclass[10pt,twocolumn,letterpaper]{article}

\usepackage{graphicx,verbatim}
\usepackage{booktabs}
\usepackage{amsmath}
\usepackage{amssymb}
\usepackage{multirow}  
\usepackage{graphicx}   
\usepackage{array}      
\usepackage{pifont}
\usepackage{colortbl}  
\usepackage{xcolor}
\usepackage{overpic}
\usepackage{tikz}
\usepackage{flushend}
\usepackage{pifont}
\usepackage{tabularx}
\usepackage{csquotes}
\usepackage{wacv}              

\usepackage{multirow}

\usepackage{xurl}

\definecolor{wacvblue}{rgb}{0.21,0.49,0.74}
\usepackage[pagebackref,breaklinks,colorlinks,allcolors=wacvblue]{hyperref}

\def\wacvPaperID{2415} 
\def\confName{WACV}
\def\confYear{2027}

\title{MUMINS: Metadata-conditioned Uncertainty-aware \\ Medical Image Next-state Synthesis}

\author{
Anna Oliveras$^{1,2}$,
Roger Mar\'i$^1$,
Rafael Redondo$^1$,
Oriol Guardi\`a-Olivella$^1$,
Cynthia Ifeyinwa Ugwu$^1$, \\
Ana Tost$^1$, 
Bhalaji Nagarajan$^3$,
Carolina Migliorelli$^1$,
Vicent Ribas$^1$,
Petia Radeva$^2$\\ [2pt]
$^1$Eurecat, Centre Tecnol\`ogic de Catalunya, Barcelona, Spain\\
$^2$Dept. de Matem\`atiques i Inform\`atica, Universitat de Barcelona, Barcelona, Spain\\
$^3$Barcelona Supercomputing Center (BSC), Barcelona, Spain\\
\\[2pt]
}

\begin{document}
\maketitle
\begin{abstract}
Forecasting anatomical changes such as tumor growth and neurodegeneration is a challenging generative vision task. Morphological evolution is subtle relative to static anatomy, highly patient-specific, and inherently stochastic. Existing methods struggle with several issues: deterministic networks ignore biological stochasticity, while standard diffusion models require computationally prohibitive multi-pass sampling to quantify uncertainty. We propose \textbf{MUMINS} (Metadata-conditioned Uncertainty-aware Medical Image Next-state Synthesis), an efficient diffusion framework that jointly diffuses a baseline scan and its follow-up residual, summed to synthesize the follow-up scan, while concurrently predicting a spatial uncertainty map, in a single reverse diffusion process. Conditioned on the time interval and relevant metadata, it preserves fine-grained anatomy by dynamically re-injecting the baseline as a soft anchor at every denoising step, and a negative-log-likelihood head learns the uncertainty map to explicitly flag error-prone regions. Designed without organ-specific heuristics, the same architecture is reused across anatomies via separate, dataset-specific retraining. Extensive evaluations demonstrate that dataset-specific retraining of MUMINS matches or outperforms dedicated, domain-specific state-of-the-art methods on lung CT (PNG) and brain MRI (OASIS-3). Project page: \textcolor{blue}{\href{https://aolivtous.github.io/publications/mumins.html}{aolivtous.github.io/publications/mumins.html}}.
\end{abstract}

\begin{figure}[t]
    \centering
    \includegraphics[trim=0 223 512 0, clip, width=0.49\textwidth]{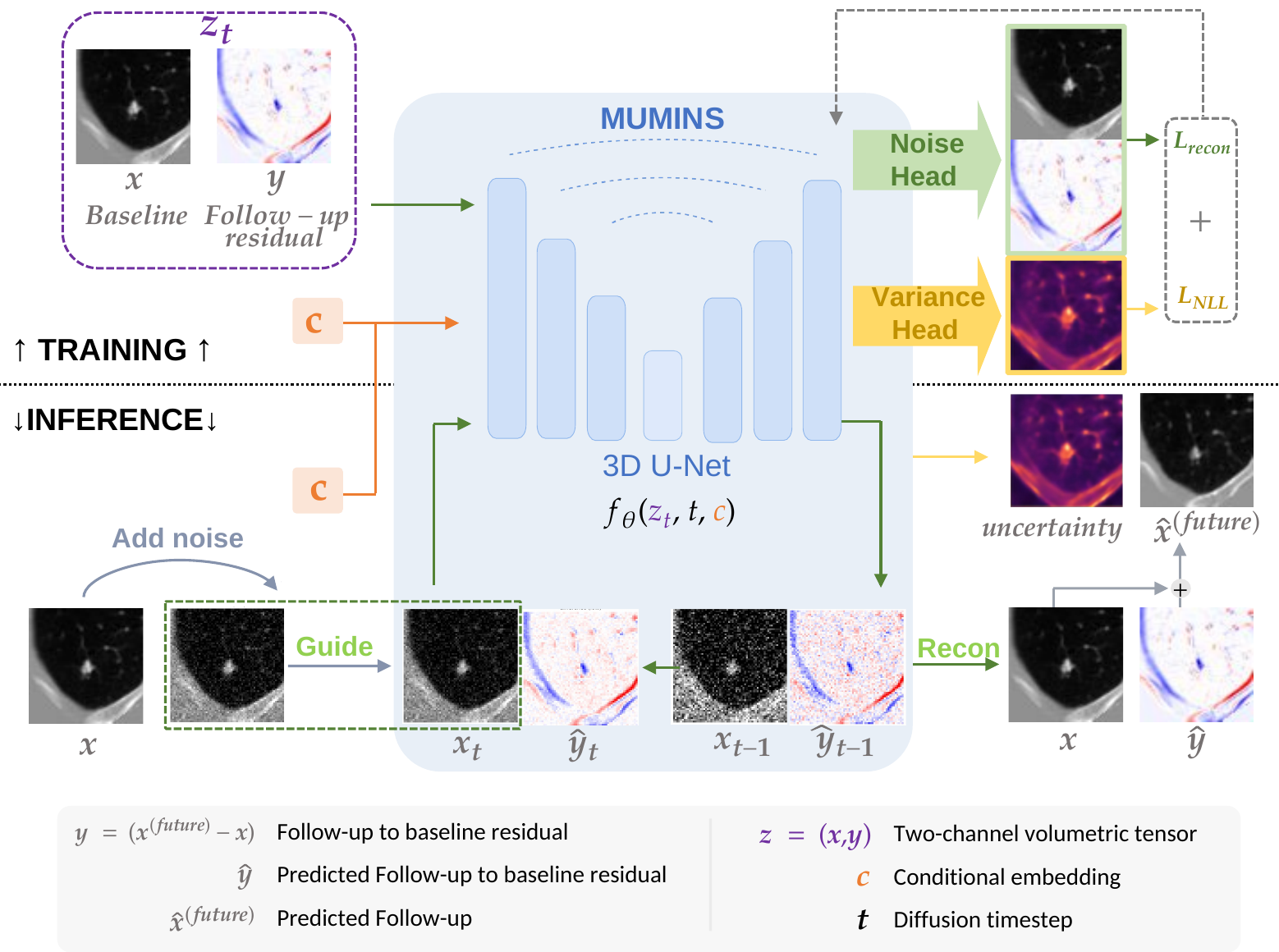}
    \caption{\!MUMINS overview. Training: the model learns to jointly generate the baseline image, the baseline-to-follow-up residual, and a voxel-wise uncertainty map via denoising diffusion. Inference: the real baseline is re-injected at every denoising step to guide residual generation. Joint state (baseline-residual) representation enhances consistency between scans, while predicted uncertainty highlights error-prone regions of the synthetic follow-up.
    }
    \label{fig:teaser}
\end{figure}

\section{Introduction}

Predicting how anatomy evolves over time is a clinical necessity. In lung cancer, Computed Tomography (CT) screening identifies millions of indeterminate pulmonary nodules yearly, and survival hinges on resolving the ``wait-and-watch'' dilemma between benign, indolent lesions and early-stage malignancies~\cite{Seijo2400889}. In neurodegeneration, longitudinal brain Magnetic Resonance Imaging (MRI) underpins the tracking of Alzheimer's disease progression, where patient-specific forecasts could accelerate trial stratification and treatment decisions~\cite{Bowles2018}. In both modalities, forecasting a future scan (follow-up) from a starting scan (baseline) at an arbitrary interval is a prognostic necessity.

From a computer-vision standpoint, this is a hard conditional generation problem: the morphological change between scans is small relative to the static anatomy, highly patient-specific, and inherently stochastic~\cite{buder2015model,hadjichrysanthou2018development}. A generative model must therefore preserve fine baseline scan details while generating plausible changes, and, since a single deterministic output cannot capture the range of biologically plausible futures, should at least flag where its own prediction is most likely to diverge from the true outcome.

Progression modeling has evolved from classical analytical models~\cite{hammer2023volume} to discriminative deep-learning predictors~\cite{NoFoNet,Tao2022,GMAE} and, recently, generative models~\cite{GPWGAN,BrLP,TADM3D,CorrFlowNet}. However, three limitations persist. First, \textit{determinism}: the dominant deterministic and kinetic-model methods yield a single future for each input. Second, \textit{organ specificity}: existing methods embed anatomy- or modality-specific components (e.g., brain-age regularizers, growth-kinetic priors), so each new setting demands a custom architecture~\cite{fan2025t,9557808}. Third, \textit{uncertainty}: recent diffusion models offer probabilistic synthesis~\cite{TADM3D,BrLP,CorrFlowNet}, but extracting a per-voxel uncertainty map requires costly Monte Carlo aggregation over many reverse-diffusion passes~\cite{wolleb2022diffusion,xie2022measurement,TaDiff}.

We propose \textbf{MUMINS} (\textbf{M}etadata-conditioned \textbf{U}ncertainty-aware \textbf{M}edical \textbf{I}mage \textbf{N}ext-state \textbf{S}ynthesis), a diffusion framework that forecasts a patient's follow-up scan (next state) at an arbitrary future interval $\Delta\tau$ from their current one (baseline), along with a spatial uncertainty map, addressing all three limitations above. Fig.~\ref{fig:teaser} gives an overview of the method. Our contributions are:

\begin{itemize}
    \item \textbf{Joint baseline-residual representation.} 
    MUMINS jointly denoises the baseline and baseline-to-follow-up residual, with baseline re-injection at each inference step to preserve patient-specific anatomy while synthesizing plausible morphological change.
    
    \item \textbf{Single-pass per-voxel uncertainty prediction.} 
    We embed likelihood-based uncertainty learning to predict voxel-wise uncertainty from a single reverse-diffusion process. This bypasses the prohibitive cost of Monte Carlo sampling, which requires multiple inferences per sample, yielding uncertainty maps that correlate with reconstruction error.
    
    \item \textbf{Domain-agnostic design.} MUMINS uses a universal architecture with generic metadata conditioning, free of organ- or modality-specific components. After dataset-specific retraining, it matches or outperforms domain-specific state-of-the-art methods on PNG lung CT~\cite{NGP} and OASIS-3 brain MRI~\cite{OASIS3}.
\end{itemize}

\section{Related Work}

Longitudinal medical image progression aims to forecast patient-specific anatomical change from baseline scans, with growing attention to the uncertainty associated with such forecasts~\cite{zou2023review}. In line with the experimental benchmarks considered in this work, we first discuss existing methods for pulmonary nodule evolution and neurodegenerative brain progression, then review recent diffusion-based approaches to uncertainty quantification.

\subsection{Longitudinal Medical Image Progression}

{\noindent \bf{Deterministic and registration-based methods.}}
Early progression models are deterministic regressors that learn a fixed mapping from baseline to follow-up. For lung nodules, NoFoNet~\cite{NoFoNet} learns voxel-wise displacement fields to warp the baseline scan, refining intensities via a dedicated TextureNet branch, while Tao et al.~\cite{Tao2022} similarly warp a prior scan conditioned on the follow-up interval. LCTformer~\cite{LCTformer} instead couples a longitudinal Transformer with a ConvLSTM to capture spatiotemporal growth dynamics. A parallel line of work couples clinical kinetic growth models (e.g., Gompertz functions) with deep generative networks: GM-AE~\cite{GMAE}, LNGNet~\cite{LNGNet}, and STGNet~\cite{STGNet} all follow a two-stage paradigm, first predicting future nodule mass or volume via kinetic equations, then conditioning a downstream decoder to synthesize the final shape and texture. Recently, NGP-Net~\cite{NGP} introduced a lightweight growth architecture and released the longitudinal Pulmonary Nodule Growth (PNG) dataset, which, unlike fixed-interval datasets such as NLST~\cite{NLST}, features irregular scan intervals handled via a dedicated spatiotemporal encoding module. Despite their differing mechanisms, deterministic approaches yield only a single predicted future without progression uncertainty, and kinetic-model variants are additionally vulnerable to error propagation, as downstream shape and texture generation depend on the accuracy of upstream growth predictions.

{\noindent \bf{Adversarial and Flow Matching methods.}}
GP-WGAN \cite{GPWGAN} proposed adversarial training for nodule progression by generating one-year follow-up scans, demonstrating that synthesized images could improve malignancy classification. To model continuous dynamics, CorrFlowNet~\cite{CorrFlowNet} maps baseline CT embeddings to follow-up embeddings via a latent Neural ODE trained with flow matching, using a correlational autoencoder to isolate nodule progression from static lung anatomy. Similarly, IMMFM~\cite{IMMFM} employs piecewise-quadratic interpolation paths as smooth targets for flow matching in longitudinal neuroimaging. While these methods can model progression over continuous time, they mainly learn transitions from one state to another rather than capturing dense distributions of plausible follow-up scans, motivating diffusion-based approaches.

{\noindent \bf{Diffusion-based progression.}} Denoising diffusion probabilistic models (DDPMs) have become a dominant paradigm for high-fidelity image generation~\cite{dhariwal2021diffusion, muller2023multimodal, pinaya2022brain, oliveras2026anatomically}. Recent methods leverage DDPMs for follow-up synthesis, but remain deeply anatomy-, disease-, or modality-specific. In pulmonary imaging, CIResDiff~\cite{CIResDiff} predicts idiopathic pulmonary fibrosis progression by learning residual differences between scans, conditioned on Contrastive Language-Image Pretraining (CLIP)~\cite{clip} embeddings. For brain MRI, TADM-3D~\cite{TADM3D} models volumetric progression as residual diffusion over intensity differences, enforcing temporal coherence via Brain-Age Estimator conditioning and Back-In-Time Regularization, while TaDiff~\cite{TaDiff} integrates treatment awareness into the diffusion process for diffuse glioma growth prediction, guided by sequential multi-parametric MRI and treatment plans. BrLP~\cite{BrLP} operates in a compact latent space using a latent diffusion model, a ControlNet~\cite{ControlNet}, an auxiliary volumetric network, and a patch discriminator for adversarial training, averaging multiple generation paths to improve spatiotemporal consistency and volumetric accuracy in Alzheimer's disease.

\subsection{Uncertainty Quantification in Diffusion Models}
Quantifying uncertainty in diffusion-based synthesis is an active, clinically relevant area. The most direct approach is Monte Carlo (MC) sampling: running the reverse process $K$ times and taking the per-voxel variance~\cite{wolleb2022diffusion, xie2022measurement, TaDiff, BrLP}. This scales as $K \times T$ network evaluations per volume and is sensitive to $K$~\cite{shaw2025bayesian}. BayesDiff~\cite{BayesDiff} reduces this cost via a post-hoc Last-Layer Laplace Approximation but still resamples weights every step, leaving the multi-pass bottleneck intact. A complementary line, from sports trajectory forecasting, instead learns uncertainty in a single forward pass: U2Diff~\cite{U2Diff} and its heteroscedastic extension U2Diffine~\cite{U2Diffine} add a bivariate NLL term to the DDPM objective, jointly predicting the noise's mean and variance. U2Diffine propagates this variance to the output via a first-order Taylor expansion of the reverse update, linearizing the denoiser via its Jacobian; U2Diff forgoes this for speed.
\section{Methodology}

In this section, we describe MUMINS in four stages. First, follow-up synthesis is formulated as conditional generation of a joint baseline-residual state, separating static anatomy from longitudinal change. Second, a 3D temporally conditioned diffusion model learns this joint state and its associated uncertainty. Third, during sampling, the baseline scan is re-injected at every reverse step to anchor the generated residual to patient-specific anatomy. Finally, uncertainty estimates from the reverse steps are accumulated into a voxel-wise map of where the synthesized follow-up is less reliable. Fig.~\ref{fig:teaser} gives an overview of the method.

\subsection{Joint Baseline-Residual Representation}
\label{sec:joint_state}
Let a patient's baseline anatomical state be an observation $\mathbf{x} \in \mathcal{X} \subset \mathbb{R}^{D \times H \times W}$ acquired at time $\tau_0$, and the follow-up state be $\mathbf{x}^{(\mathrm{future})}$ acquired at $\tau_{\mathrm{future}}$, with temporal interval $\Delta\tau = \tau_{\mathrm{future}} - \tau_0$.
Directly modeling $p(\mathbf{x}^{(\mathrm{future})} \mid \mathbf{x}, \Delta\tau)$ often leads to shortcut learning, where the network ignores the temporal condition or fails to preserve patient-specific anatomy~\cite{TADM3D, TaDiff}. We instead target the longitudinal residual $\mathbf{y} = \mathbf{x}^{(\mathrm{future})} - \mathbf{x}$, isolating disease progression and anatomical evolution from the static anatomy, and synthesize it jointly with the baseline as $\mathbf{z} = (\mathbf{x}, \mathbf{y})$.
This formulation is inspired by DiffAtlas~\cite{DiffAtlas}, a segmentation method that jointly diffuses an image and its mask and, at inference, substitutes the known image at each reverse step to guide mask generation. We adapt this idea to progression along two axes. First, rather than an image-mask pair, our joint state pairs the static baseline $\mathbf{x}$ with the dynamic residual $\mathbf{y}$, so that substituting the known baseline at inference (Sec.~\ref{sec:reinj}) guides the generated residual. Second, we condition on clinical and acquisition variables $\mathbf{c}$ (Sec.~\ref{sec:conditioning}) to further encourage patient- and time-specific synthesis. We therefore learn the conditional joint distribution $p_\theta(\mathbf{z} \mid \mathbf{c})$; modeling baseline and residual jointly keeps the generated progression structurally consistent with the baseline.
\subsection{Forward Process and Noise Prediction}
\label{sec:forward}
We construct a discrete-time Markovian diffusion process that progressively adds Gaussian noise to the joint data distribution $\mathbf{z}_0 \sim p_{\mathrm{data}}(\mathbf{z})$~\cite{ho2020denoising}. Following~\cite{nichol2021improved}, the noise is scheduled with a cosine variance profile $\beta_1, \dots, \beta_T$ over $T$ steps, defining the forward transitions as $q(\mathbf{z}_t \mid \mathbf{z}_{t-1}) = \mathcal{N}(\mathbf{z}_t; \sqrt{1 - \beta_t}\, \mathbf{z}_{t-1}, \beta_t \mathbf{I})$. Using the reparameterization $\bar{\alpha}_t = \prod_{s=1}^t (1 - \beta_s)$, the marginal distribution at any diffusion timestep $t$ is given in closed form:
\begin{equation}
q(\mathbf{z}_t \mid \mathbf{z}_0) = \mathcal{N}(\mathbf{z}_t; \sqrt{\bar{\alpha}_t}\, \mathbf{z}_0, (1 - \bar{\alpha}_t)\mathbf{I}).
\end{equation}
The reverse process is parameterized by a 3D U-Net~\cite{cciccek20163d} $f_\theta(\mathbf{z}_t, t, \mathbf{c})$ outputting four channels, split into a noise-prediction tensor $\hat{\boldsymbol\epsilon}_\theta(\mathbf z_t,t,\mathbf c)$ (the noise-mean head) and a logit tensor $\boldsymbol{\ell}_\theta(\mathbf{z}_t, t, \mathbf{c})$ passed through a sigmoid to produce a per-voxel standard deviation (the variance head):
\begin{equation}
\hat{\boldsymbol{\sigma}}_\theta(\mathbf{z}_t, t, \mathbf{c}) = \mathrm{sigmoid}(\boldsymbol{\ell}_\theta(\mathbf{z}_t, t, \mathbf{c})) + \varepsilon,
\label{eq:sigma_head}
\end{equation}
with $\varepsilon = 10^{-6}$ for numerical stability. Because $\mathbf{z}_t$ is a joint variable, both outputs factorize into baseline and residual components, $\hat{\boldsymbol{\epsilon}}_\theta = (\hat{\boldsymbol{\epsilon}}^x_\theta, \hat{\boldsymbol{\epsilon}}^y_\theta)$ and $\hat{\boldsymbol{\sigma}}_\theta = (\hat{\boldsymbol{\sigma}}^x_\theta, \hat{\boldsymbol{\sigma}}^y_\theta)$.
For brevity, we henceforth write $\hat{\boldsymbol{\epsilon}}_\theta \equiv \hat{\boldsymbol{\epsilon}}_\theta(\mathbf{z}_t, t, \mathbf{c})$ and $\hat{\boldsymbol{\sigma}}_\theta \equiv \hat{\boldsymbol{\sigma}}_\theta(\mathbf{z}_t, t, \mathbf{c})$, with the same convention applied to their per-channel components $\hat{\boldsymbol{\epsilon}}^{k}_\theta$ and $\hat{\boldsymbol{\sigma}}^{k}_\theta$ for $k \in \{x, y\}$, suppressing dependence on $(\mathbf{z}_t, t, \mathbf{c})$ unless required for clarity.

\subsection{Temporal and Metadata Conditioning}
\label{sec:conditioning}
The conditioning vector $\mathbf{c} = (\Delta \tau, a, \Delta b_x, \Delta b_y, \Delta b_z)$ aggregates clinical and acquisition signals relevant to longitudinal modeling: the inter-scan interval $\Delta \tau$, the patient's age at $\tau_0$ ($a$), and a triplet of acquisition-quality descriptors $(\Delta b_x, \Delta b_y, \Delta b_z)$ encoding the per-axis change in acquisition sharpness between scans, $\Delta b_i = b_i(\mathbf{x}^{(\mathrm{future})}) - b_i(\mathbf{x})$, where $b_i$ is the Tenengrad sharpness (mean squared image gradient along axis $i$), so positive values indicate a sharper, less blurred follow-up. At inference, we report two settings: $\Delta\mathbf{b} = \mathbf{0}$, which preserves the baseline's own sharpness and requires no knowledge of the unseen follow-up, and $\Delta\mathbf{b} = \mathrm{GT}$, an oracle setting computed directly from the true follow-up. The former is the only information-respecting choice in a genuine forecasting deployment; the latter is reported purely as a diagnostic (Table~\ref{tab:results}). Follow-up scans can otherwise exhibit global sharpness differences from acquisition variability rather than anatomical progression, as in the PNG benchmark~\cite{NGP}; blur conditioning disentangles these acquisition-related appearance changes from true anatomical change (See Supp. Sec. G).
Each scalar is independently mapped through a sinusoidal positional embedding~\cite{vaswani2017attention} and a two-layer MLP, then concatenated with the diffusion timestep embedding. This combined embedding modulates the network's intermediate representations via the time-conditioned residual blocks, letting the noise estimator adapt its transition dynamics to the current noise scale, target longitudinal horizon, and acquisition context.

 \subsection{Hybrid Objective with Voxel-wise Uncertainty}
\label{sec:hybrid}
We adapt the NLL-based uncertainty learning of U2Diff~\cite{U2Diff}, originally developed for trajectory forecasting, to volumetric follow-up synthesis. The model predicts a per-voxel variance alongside each diffusion noise estimate, enabling uncertainty maps to be obtained from a single reverse-diffusion process. Training uses a hybrid objective that combines standard noise prediction with the NLL term.

{\noindent \bf{Joint Noise Prediction.}} The primary objective trains the network to denoise the joint state. For each sample, we draw $\boldsymbol{\epsilon} \sim \mathcal{N}(\mathbf{0}, \mathbf{I})$ and $t \sim \mathcal{U}(1, T)$, and minimize a channel-wise reconstruction loss on the predicted noise mean:
\begin{equation}
\mathcal{L}_{\mathrm{recon}} = \mathbb{E}_{\mathbf{z}_0, \boldsymbol{\epsilon}, t, \mathbf{c}}
\frac{1}{N}\left[ \left\| \boldsymbol{\epsilon}^x - \hat{\boldsymbol{\epsilon}}^x_\theta \right\|_1
+ \left\| \boldsymbol{\epsilon}^y - \hat{\boldsymbol{\epsilon}}^y_\theta \right\|_1 \right],
\label{eq:recon}
\end{equation}
where $N = DHW$ is the number of voxels; $x$ and $y$ channels are weighted equally so that neither dominates. 

{\noindent \bf{Uncertainty Loss.}} To learn predictive variances without destabilizing noise prediction, we add an NLL term where the predicted mean acts only as a target for the variance head (Sec.~\ref{sec:forward}), whose per-voxel standard deviation $\hat{\boldsymbol{\sigma}}^{k}_\theta$ (Eq.~\ref{eq:sigma_head}) is the predicted noise scale in the Gaussian NLL of Eq.~\eqref{eq:nll}. Following~\cite{U2Diff}, we apply a stop-gradient $\mathrm{sg}[\cdot]$ to the noise mean so gradients from the NLL flow only through the uncertainty head:
\begin{equation}
\begin{aligned}
\mathcal{L}_{\mathrm{NLL}} = \mathbb{E}_{\mathbf{z}_0, \boldsymbol{\epsilon}, t, \mathbf{c}}
\frac{1}{N}\sum_{k \in \{x, y\}} \sum_{i=1}^{N} \bigg[ &\log\!\left(\sqrt{2\pi}\,\hat{\sigma}^{k}_{\theta,i}\right) \\
&+ \frac{\left(\epsilon^k_i - \mathrm{sg}\!\left[\hat{\epsilon}^{k}_{\theta,i}\right]\right)^2}
{2\,(\hat{\sigma}^{k}_{\theta,i})^2} \bigg],
\end{aligned}
\label{eq:nll}
\end{equation}
This decomposition decouples optimizing the deterministic noise estimate (Eq.~\ref{eq:recon}) from calibrating its uncertainty, preventing the network from trivially reducing $\mathcal{L}_{\mathrm{NLL}}$ by inflating $\hat{\boldsymbol{\sigma}}_\theta$ at the cost of $\mathcal{L}_{\mathrm{recon}}$. The total objective is:
\begin{equation}
\mathcal{L}_{\mathrm{total}} = \mathcal{L}_{\mathrm{recon}} + \lambda_{\mathrm{NLL}} \, \mathcal{L}_{\mathrm{NLL}},
\end{equation}
with $\lambda_{\mathrm{NLL}} = 0.01$ set empirically so the uncertainty term does not dominate denoising.
\subsection{Baseline-Guided Conditional Synthesis}
\label{sec:reinj}
During inference, we sample from the conditional distribution $p_\theta(\mathbf{x}, \mathbf{y} \mid \mathbf{c})$ given an observed baseline $\mathbf{x}$ and the conditioning vector $\mathbf{c}$.
Because our model is trained on this joint distribution, we perform conditional sampling via baseline injection. Starting from pure Gaussian noise $\mathbf{z}_T \sim \mathcal{N}(\mathbf{0}, \mathbf{I})$, at each reverse step $t$ the network predicts the joint posterior parameters. To ground the generative process in the patient's actual anatomy, prior to each network evaluation we overwrite the $\mathbf{x}$-component of $\mathbf{z}_t$ with an exact sample from the forward marginal of the known baseline:
\begin{equation}
\mathbf{x}_t \sim \mathcal{N}\!\left(\sqrt{\bar\alpha_t}\, \mathbf{x},\ (1-\bar\alpha_t)\mathbf{I}\right).
\label{eq:guide}
\end{equation}
We draw a single noise realization $\bar{\boldsymbol{\epsilon}} \sim \mathcal{N}(\mathbf{0}, \mathbf{I})$ per sample and reuse it at every reverse step, i.e.\ $\mathbf{x}_t = \sqrt{\bar\alpha_t}\,\mathbf{x} + \sqrt{1-\bar\alpha_t}\,\bar{\boldsymbol{\epsilon}}$ for all $t$. This satisfies the marginal of Eq.~\eqref{eq:guide} at every step while tracing one coherent forward trajectory of the baseline, so the residual $\mathbf{y}_t$ is guided by temporally consistent anatomical context; it outperforms per-step noise resampling (Supp.~Sec. F, Table F.5). The reverse process yields  the denoised joint state $\hat{\mathbf{z}}_0 = (\hat{\mathbf{x}}_0, \hat{\mathbf{y}}_0)$, and the synthesized follow-up is $\hat{\mathbf{x}}^{(\mathrm{future})} = \mathbf{x} + \hat{\mathbf{y}}_0$.

\subsection{Uncertainty Propagation at Inference}
\label{sec:variance_prop}
Beyond a single volume, our framework yields a per-voxel uncertainty map by propagating the network's predicted variance through the reverse process, under two approximations that make this tractable at volumetric resolution: a diagonal predictive covariance, so the network outputs a per-voxel variance $\hat{\boldsymbol{\sigma}}_\theta^2$; and, as in U2Diff~\cite{U2Diff}, a zero Jacobian $\mathbf{J}_t = \nabla_{\mathbf{z}} \hat{\boldsymbol{\epsilon}}_\theta$, since retaining it requires a backward pass through the 3D denoiser at every reverse step, infeasible at our scale even under the diagonal approximation~\cite{U2Diffine}. Our sampling uses the stochastic DDIM~\cite{DDIM} update with skip interval $\zeta$ and stochasticity $\eta$, with per-step noise scale $\sigma_t$. Let $\mathbf{V}_t = \operatorname{Var}(\mathbf{z}_t)$ be the accumulated per-voxel variance ($\mathbf{V}_T = \mathbf{0}$). Modeling the network's predictive uncertainty as a zero-mean term $\boldsymbol\xi_t$ with variance $\hat{\boldsymbol\sigma}_\theta^2$, independent of $\mathbf z_t$ and of the injected noise, a first-order variance decomposition yields:
\begin{equation}
\label{eq:var_recursion}
\mathbf{V}_{t-\zeta} = r_t^2\, \mathbf{V}_t + c_t^2\, \hat{\boldsymbol{\sigma}}_\theta^2 + \sigma_t^2,
\end{equation}
with $r_t = \sqrt{\bar{\alpha}_{t-\zeta}/\bar{\alpha}_t}$ and $c_t = \sqrt{1 - \bar{\alpha}_{t-\zeta} - \sigma_t^2} - r_t\sqrt{1 - \bar{\alpha}_t}$. The three terms propagate accumulated uncertainty along the trajectory, add the network's predicted uncertainty at step $t$, and inject the sampler's stochastic noise, respectively; the full derivation is in Supp.~Sec. A. Setting $\mathbf{J}_t = \mathbf{0}$ removes the stabilizing cross-term of the full formulation~\cite{U2Diffine}, leaving an unopposed map with per-step amplification $r_t^2 > 1$, so propagating from the start of the chain accumulates variance monotonically into over-dispersed, poorly calibrated maps (a structural, not numerical, effect; Supp.~Sec. B). We therefore adopt the delayed variance propagation of U2Diff~\cite{U2Diff}, holding $\mathbf{V}_t=\mathbf{0}$ for $t\ge\hat s$ and applying the recursion from step $\hat s$ onward, selecting $\hat s$ by minimizing the NLL. Unlike U2Diff and U2Diffine, which sample deterministically, our baseline-anchored synthesis requires \emph{stochastic} sampling: the injected term $\sigma_t^2$ enters the recursion only for $\eta>0$, and is needed both to calibrate uncertainty and, as our ablation shows, to maintain reconstruction quality (Supp.~Sec. F, Table F.4).

\paragraph{From residual variance to follow-up uncertainty.}
The recursion above propagates variance over the two-channel joint state, so the final map $\mathbf{V}_0$ contains a baseline component $\mathbf{V}_0^{x}$ and a residual component $\mathbf{V}_0^{y}$. Since the synthesized follow-up is $\hat{\mathbf{x}}^{(\mathrm{future})} = \mathbf{x} + \hat{\mathbf{y}}_0$ with the baseline $\mathbf{x}$ injected as a known measurement, its predictive variance reduces exactly to the residual component:
\begin{equation}
\label{eq:varCal}
\operatorname{Var}\!\big(\hat{\mathbf{x}}^{(\mathrm{future})}\big)
  = \operatorname{Var}(\mathbf{x} + \hat{\mathbf{y}}_0)
  = \operatorname{Var}(\hat{\mathbf{y}}_0)
  = \mathbf{V}_0^{y},
\end{equation}
where $\operatorname{Var}(\mathbf{x})=0$ and $\operatorname{Cov}(\mathbf{x},\hat{\mathbf{y}}_0)=0$ because $\mathbf{x}$ is deterministic and fully observed. This holds under either guide-noise scheme (Sec.~\ref{sec:reinj}), so the decomposition of Eq.~\eqref{eq:var_recursion} is unaffected by the single-trajectory-vs-resampling choice. $\mathbf{V}_0^{y}$ is thus the per-voxel predictive variance of the synthesized follow-up, and we report its element-wise square root as the final uncertainty map, $\boldsymbol{\sigma} \triangleq \sqrt{\mathbf{V}_0^{y}}$, distinct from the per-step network output $\hat{\boldsymbol{\sigma}}_\theta$ (Eq.~\ref{eq:sigma_head}) and the accumulated variance $\mathbf{V}_t$ of Eq.~\eqref{eq:var_recursion}. As in U2Diff~\cite{U2Diff}, the DDIM grid ends at $t_1=1$, and a final standard DDPM~\cite{ho2020denoising} step to $t=0$ carries the accumulated variance unchanged, $\mathbf{V}_0=\mathbf{V}_{t_1}$.

\section{Experiments and Results}
\subsection{Datasets and Preprocessing}
\label{sec:datasets}

\noindent\textbf{Lung Nodule Progression.} We use the Pulmonary Nodule Growth (PNG) dataset~\cite{NGP}: 378 chest CT scans from 103 patients, 226 longitudinal nodules with at least three time points (2--64 month intervals), radiologist-verified. We follow NGP's~\cite{NGP} preprocessing and splits, converting longitudinal trios into pairs while keeping patients within a single split (full pipeline in Supp.~Sec. C).

\noindent\textbf{Alzheimer's Disease Progression.} We use OASIS-3~\cite{OASIS3}: 1352 T1-weighted scans from 492 subjects (ages 42--97; CN/MCI/AD), spanning $\sim$10 years with a mean baseline-to-follow-up interval of 3.7 years. We follow the Turboprep~\cite{puglisi2024turboprep} preprocessing of~\cite{BrLP} and use $128^3$ volumes as in~\cite{TADM3D}, split $70/15/15\%$ with intra-patient pairs kept within a single split (full pipeline in Supp.~Sec. C).

\subsection{Implementation Details}
MUMINS uses a 3D U-Net~\cite{cciccek20163d} with per-resolution attention factorized into in-plane and rotary-embedding depth-axis terms~\cite{su2024roformer}, adapted from the spatial/temporal split used in video diffusion models. We train with Adam ($10^{-4}$) and EMA (decay 0.995) for 150k iterations, batch 2 (300k for brain at batch 1; Supp. D) on 2 NVIDIA H100 (64\,GB) GPUs, using $T = 300$ diffusion steps with a cosine schedule~\cite{nichol2021improved} and $\lambda_{\mathrm{NLL}} = 0.01$. At inference we use stochastic DDIM~\cite{DDIM} with skip interval $\zeta=5$ ($60$ steps), $\eta=1$, and variance start-step $\hat{s}=15$ (selected on held-out validation; ablated in Supp. F). All baselines are retrained and evaluated on identical splits; extended details in Supp. D.

\begin{figure*}[tp]
    \centering
    \includegraphics[width=0.73\textwidth]{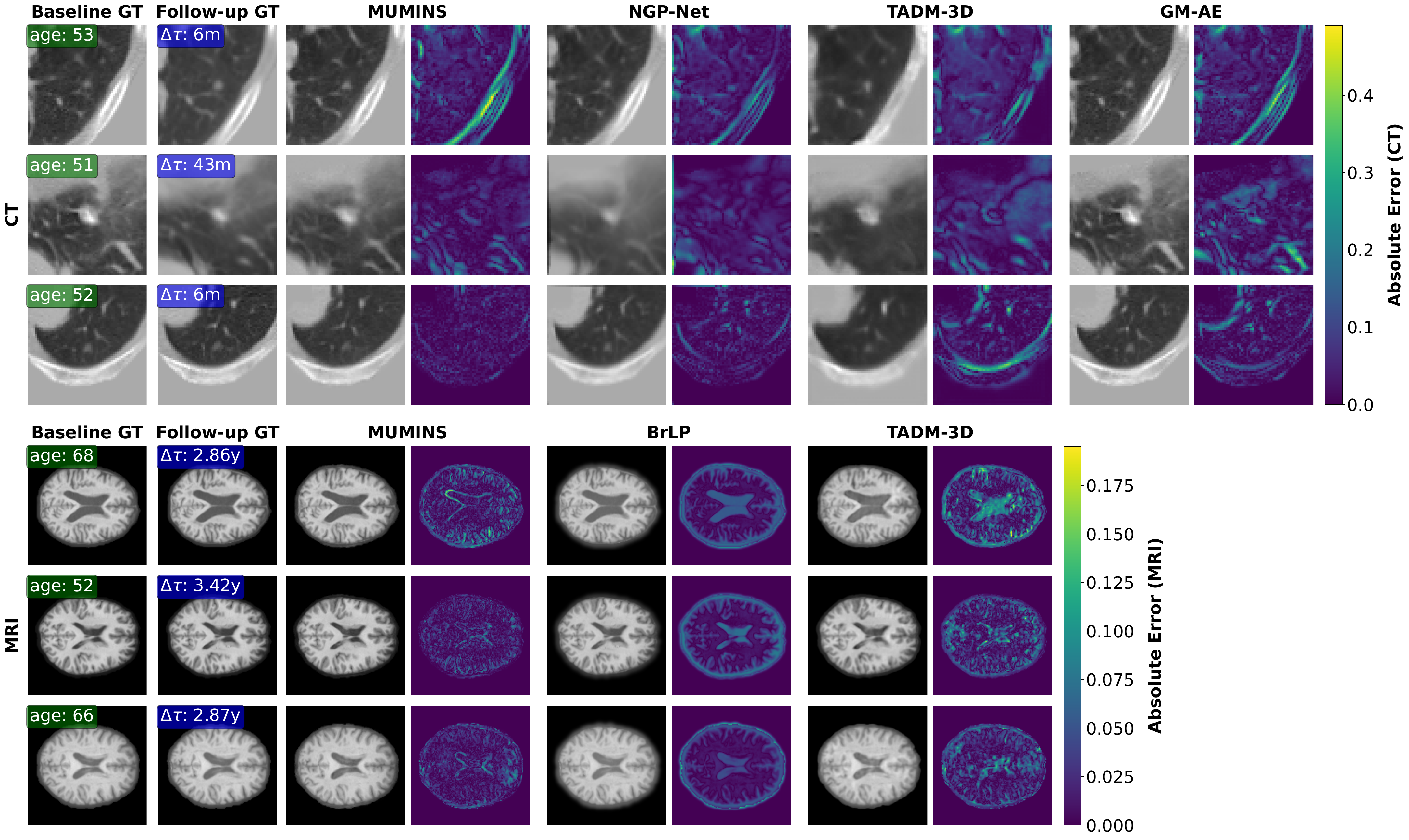} 
  \caption{Qualitative next-state synthesis on lung CT (top) and brain MRI (bottom): baseline (age, green), ground-truth follow-up ($\Delta \tau$, blue; in years \texttt{y} or months \texttt{m}), and per-method predictions with absolute error maps. MUMINS inference with $\Delta\mathbf{b} = \mathbf{0}$. }
    \label{fig:NGPReuslts}
\end{figure*}
\subsection{Evaluation metrics}
\label{sec:metrics}
As described in Sec.~\ref{sec:datasets}, samples are organized as longitudinal pairs; all metrics are averaged over pairs rather than subjects or nodules.

\noindent\textbf{Synthesis quality.} 
We compare each generated follow-up $\hat{\mathbf{x}}^{(\mathrm{future})}$ against the ground truth $\mathbf{x}^{(\mathrm{future})}$ using three complementary metrics: mean absolute error (MAE\,$\downarrow$), the average voxel-wise intensity error; peak signal-to-noise ratio (PSNR\,$\uparrow$), the reconstruction quality to the maximum signal; and structural similarity (SSIM\,$\uparrow$)~\cite{wang2004image}, the perceptual and structural agreement beyond intensity.

\noindent\textbf{Uncertainty evaluation.} MUMINS outputs a per-voxel uncertainty map $\boldsymbol{\sigma}$ in a single reverse-diffusion pass. We assess $\boldsymbol{\sigma}$ against the absolute error map $\mathbf{e}$, with $e_i = |x^{(\mathrm{future})}_i - \hat{x}^{(\mathrm{future})}_i|$, along two axes: \emph{localization}, whether $\boldsymbol{\sigma}$ is high where error is large, and \emph{calibration}, whether the predicted uncertainty magnitude matches the observed error scale. For \emph{localization}, we report Spearman ($\rho$) correlation between $\boldsymbol{\sigma}$ and $\mathbf{e}$, and the Area Under the Sparsification Error curve (AUSE); lower is better. For \emph{calibration}, we report the $95\%$ prediction-interval coverage probability ($\mathrm{PICP}_{95}$, ideal $0.95$) and the mean prediction-interval width (MPIW), computed in normalized image-intensity units (Supp.~Sec. D). Coverage at additional nominal levels is given in Supp.~Sec. E. As a widely used uncertainty estimate in diffusion models~\cite{wolleb2022diffusion,xie2022measurement,TaDiff,BrLP}, we compute a Monte Carlo (MC) reference: for each input we draw $K=20$ stochastic generations from the same trained model and take their voxel-wise standard deviation. Since MC and our single-pass $\boldsymbol{\sigma}$ capture different uncertainty sources, we do not treat MC as ground truth; both are scored independently against the true error, and we report their cross-agreement ($\rho$). We discuss the compromise of this $K{=}20$ choice, including MC's own convergence behavior, in Supp.~Sec. E.

\subsection{Results}
\label{sec:unc_results}

\begin{table*}[tp]
\centering
\footnotesize
\setlength{\tabcolsep}{3.5pt}
\caption{Synthesis results on the PNG lung CT benchmark and OASIS-3 brain MRI; subgroups indicate nodule growth trend (lung) and cognitive status (brain). Metrics reported over the whole image and the ROI (nodule/brain). \textit{Copy} is a no-change reference.}
\label{tab:results}
\begin{tabular}{c ll ccc cc}
\toprule
& \textbf{Subgroup} & \textbf{Method} & \textbf{MAE ($\times 10^{-2}$) $\downarrow$} & \textbf{PSNR (dB) $\uparrow$} & \textbf{SSIM (\%) $\uparrow$} & \textbf{MAE$_{\text{ROI}}$ ($\times 10^{-2}$) $\downarrow$} & \textbf{PSNR$_{\text{ROI}}$ (dB) $\uparrow$} \\
\midrule
\multirow{21}{*}{\rotatebox{90}{\textit{\textbf{PNG Lung CT~\cite{NGP}}}}} 
& \multirow{5}{*}{\textbf{Grow}}
    & GM-AE~\cite{GMAE}    & 4.03 $\pm$ 1.32 & 22.83 $\pm$ 2.61 & 65.29 $\pm$ 11.39 & 4.47 $\pm$ 2.51 & 25.18 $\pm$ 4.11 \\
    && TADM-3D~\cite{TADM3D}  & 4.93 $\pm$ 1.57 & 21.49 $\pm$ 2.41 & 68.33 $\pm$ 10.10 & 5.92 $\pm$ 4.37 & 23.87 $\pm$ 5.28  \\
    && NGP-Net~\cite{NGP}  & 4.13 $\pm$ 1.46 & 22.84 $\pm$ 2.80 & 71.37 $\pm$ 10.39 & 4.66 $\pm$ 2.88 & 25.06 $\pm$ 4.26 \\
    && \textbf{MUMINS $\Delta\mathbf{b} = \mathbf{0}$} &  \textbf{3.87 $\pm$ 1.60} & \textbf{23.65 $\pm$ 3.87} & \textbf{71.66 $\pm$ 12.36} & \textbf{4.23 $\pm$ 2.71} & \textbf{26.12 $\pm$ 4.81} \\
    && \textit{\textbf{MUMINS} $\Delta\mathbf{b} = \mathrm{GT}$} & \textit{2.95 $\pm$ 1.29} & \textit{26.22 $\pm$ 4.41} & \textit{80.82 $\pm$ 11.21} & \textit{3.42 $\pm$ 2.16} & \textit{28.39 $\pm$ 5.29 }  \\
\cmidrule{2-8}
& \multirow{5}{*}{\textbf{Stable}}
    & GM-AE    & \textbf{3.71 $\pm$ 1.45} & 23.75 $\pm$ 2.93 & 67.34 $\pm$ 15.02 & \textbf{7.02 $\pm$ 5.47} & 22.46 $\pm$ 6.03  \\
    && TADM-3D  & 4.46 $\pm$ 1.20 & 22.12 $\pm$ 2.14 & 67.53 $\pm$ 11.35 & 8.34 $\pm$ 6.40 & 20.91 $\pm$ 5.60 \\
    && NGP-Net  & 3.84 $\pm$ 1.64 & 23.81 $\pm$ 3.16 & 71.76 $\pm$ 14.94 & 7.11 $\pm$ 5.84 & \textbf{22.66 $\pm$ 6.22} \\
    && \textbf{MUMINS $\Delta\mathbf{b} = \mathbf{0}$} & 3.79 $\pm$ 1.90 & \textbf{24.36 $\pm$ 4.35} & \textbf{72.41 $\pm$ 15.06} & 7.52 $\pm$ 5.67 & 22.43 $\pm$ 7.43  \\
    && \textit{\textbf{MUMINS} $\Delta\mathbf{b} = \mathrm{GT}$} & \textit{3.06 $\pm$ 1.71} & \textit{26.16 $\pm$ 4.94} & \textit{76.91 $\pm$ 15.83 } & \textit{6.00 $\pm$ 5.80} & \textit{24.85 $\pm$ 7.98} \\
\cmidrule{2-8}
& \multirow{5}{*}{\textbf{Shrink}}
    & GM-AE    & 4.18 $\pm$ 1.46 & 22.67 $\pm$ 3.59 & 65.61 $\pm$ 13.55 & \textbf{5.40 $\pm$ 3.71} & \textbf{24.75 $\pm$ 5.96} \\
    && TADM-3D  & 5.03 $\pm$ 1.50 & 21.03 $\pm$ 2.72 & 64.69 $\pm$ 12.31 & 6.60 $\pm$ 4.08 & 22.24 $\pm$ 4.89  \\
    && NGP-Net  & 4.35 $\pm$ 1.53 & 22.37 $\pm$ 3.19 & 70.03 $\pm$ 12.71 & 5.69 $\pm$ 3.71 & 23.96 $\pm$ 5.57 \\
    && \textbf{MUMINS $\Delta\mathbf{b} = \mathbf{0}$} & \textbf{4.09 $\pm$ 1.75} & \textbf{23.12 $\pm$ 3.95} & \textbf{70.76 $\pm$ 14.38} & 5.73 $\pm$ 4.68 & 24.26 $\pm$ 5.75  \\
    && \textit{\textbf{MUMINS} $\Delta\mathbf{b} = \mathrm{GT}$} & \textit{3.47 $\pm$ 1.73} & \textit{24.43 $\pm$ 4.50} & \textit{75.08 $\pm$ 14.90} & \textit{4.82 $\pm$ 3.52} & \textit{25.45 $\pm$ 5.97}  \\
\cmidrule{2-8}   
& \multirow{6}{*}{\textbf{All above}}
    & \textit{Copy ($\hat{\mathbf{x}}^{(\mathrm{future})}=\mathbf{x}$)} & \textit{4.54 $\pm$ 1.56} & \textit{21.91 $\pm$ 3.07} & \textit{65.21 $\pm$ 12.66} & \textit{5.82 $\pm$ 4.02} & \textit{23.63 $\pm$ 5.21} \\
    && GM-AE  & 4.05 $\pm$ 1.38 & 22.88 $\pm$ 3.06 & 65.69 $\pm$ 12.65 & \textbf{5.17 $\pm$ 3.56} & 24.65 $\pm$ 5.19  \\
    && TADM-3D  & 4.91 $\pm$ 1.50 & 21.39 $\pm$ 2.52 & 66.77 $\pm$ 11.24 & 6.51 $\pm$ 4.60 & 22.83 $\pm$ 5.24  \\
    && NGP-Net  & 4.18 $\pm$ 1.51 & 22.78 $\pm$ 3.02 & 70.88 $\pm$ 11.93 & 5.40 $\pm$ 3.77 & 24.30 $\pm$ 5.12 \\
    \rowcolor{blue!15} \cellcolor{white}
    & \cellcolor{white} & \textbf{MUMINS $\Delta\mathbf{b} = \mathbf{0}$} & \textbf{3.95 $\pm$ 1.69} & \textbf{23.53 $\pm$ 3.96} & \textbf{71.40 $\pm$ 13.47} & 5.26 $\pm$ 4.16 & \textbf{24.89 $\pm$ 5.69} \\
    \rowcolor{blue!15} \cellcolor{white}
    & \cellcolor{white} & \textit{\textbf{MUMINS} $\Delta\mathbf{b} = \mathrm{GT}$} & \textit{3.18 $\pm$ 1.55} & \textit{25.50 $\pm$ 4.56 }& \textit{78.01 $\pm$ 13.59} & \textit{4.32 $\pm$ 3.48} & \textit{26.75 $\pm$ 6.12} \\
\midrule
\multirow{17}{*}{\rotatebox{90}{\textit{\textbf{OASIS-3 Brain MRI~\cite{OASIS3}}}}} 
& \multirow{4}{*}{\textbf{CN}}
    & BrLP~\cite{BrLP}    & 1.57 $\pm$ 0.24 & 27.16 $\pm$ 1.31 & 94.15 $\pm$ 1.19 & 6.90 $\pm$ 1.11 & 20.59 $\pm$ 1.34\\
    && TADM-3D~\cite{TADM3D}  & 1.21 $\pm$ 0.24 & 30.12 $\pm$ 1.73 & 95.00 $\pm$ 1.59 & \textbf{3.82 $\pm$ 0.80} & \textbf{26.06 $\pm$ 1.67} \\
    && \textbf{MUMINS $\Delta\mathbf{b} = \mathbf{0}$} & \textbf{1.15 $\pm$ 0.46} & \textbf{30.78 $\pm$ 2.87} & \textbf{95.21 $\pm$ 2.82} & 4.16 $\pm$ 1.84 & 25.66 $\pm$ 3.22  \\
    && \textit{\textbf{MUMINS} $\Delta\mathbf{b} = \mathrm{GT}$} & \textit{1.06 $\pm$ 0.34} & \textit{31.34 $\pm$ 2.40} & \textit{95.52 $\pm$ 2.37} & \textit{3.69 $\pm$ 1.14} & \textit{26.37 $\pm$ 2.47} \\
\cmidrule{2-8}
& \multirow{4}{*}{\textbf{MCI}}
    & BrLP    & 1.55 $\pm$ 0.20 & 27.17 $\pm$ 1.10 & 94.24 $\pm$ 1.12 & 6.88 $\pm$ 0.91 & 20.58 $\pm$ 1.09 \\
    && TADM-3D & 1.19 $\pm$ 0.32 & 30.19 $\pm$ 1.95 & 95.02 $\pm$ 1.69 & 3.69 $\pm$ 1.02 & 26.47 $\pm$ 1.92 \\
    && \textbf{MUMINS $\Delta\mathbf{b} = \mathbf{0}$} & \textbf{0.98 $\pm$ 0.33} & \textbf{31.72 $\pm$ 2.17} & \textbf{96.25 $\pm$ 1.82} & \textbf{3.42 $\pm$ 1.21} & \textbf{27.01 $\pm$ 2.39} \\
    && \textit{\textbf{MUMINS} $\Delta\mathbf{b} = \mathrm{GT}$} & \textit{0.95 $\pm$ 0.31} & \textit{31.82 $\pm$ 2.17} & \textit{96.61 $\pm$ 1.41} & \textit{3.43 $\pm$ 1.21} & \textit{27.00 $\pm$ 2.44} \\
\cmidrule{2-8}
& \multirow{4}{*}{\textbf{AD}}
    & BrLP    & 1.73 $\pm$ 0.25 & 26.17 $\pm$ 1.34 & 93.47 $\pm$ 1.36 & 7.74 $\pm$ 1.19 & 19.52 $\pm$ 1.39  \\
    && TADM-3D & 1.17 $\pm$ 0.21 & 30.01 $\pm$ 1.60 & 94.77 $\pm$ 2.15 & \textbf{3.79 $\pm$ 0.73}  & \textbf{26.01 $\pm$ 1.56} \\
    && \textbf{MUMINS $\Delta\mathbf{b} = \mathbf{0}$} & \textbf{1.16 $\pm$ 0.29} & \textbf{30.44 $\pm$ 1.97} & \textbf{95.38 $\pm$ 2.03} & 4.18 $\pm$ 1.05 & 25.35 $\pm$ 2.06  \\
    && \textit{\textbf{MUMINS} $\Delta\mathbf{b} = \mathrm{GT}$} & \textit{1.17 $\pm$ 0.31} & \textit{30.44 $\pm$ 2.16} & \textit{95.52 $\pm$ 1.74} & \textit{4.27 $\pm$ 1.21} & \textit{25.26 $\pm$ 2.31} \\
\cmidrule{2-8}
& \multirow{5}{*}{\textbf{All above}}
    & \textit{Copy ($\hat{\mathbf{x}}^{(\mathrm{future})}=\mathbf{x}$)} & \textit{1.14 $\pm$ 0.56} & \textit{30.91 $\pm$ 3.60} & \textit{95.55 $\pm$ 3.37} & \textit{4.34 $\pm$ 2.25} & \textit{25.61 $\pm$ 3.98}  \\
    && BrLP    & 1.57 $\pm$ 0.24 & 27.12 $\pm$ 1.30 & 94.12 $\pm$ 1.20 & 6.94 $\pm$ 1.10 & 20.54 $\pm$ 1.33 \\
    && TADM-3D & 1.21 $\pm$ 0.25 & 30.12 $\pm$ 1.75 & 94.99 $\pm$ 1.62 & \textbf{3.80 $\pm$ 0.82} & \textbf{26.10 $\pm$ 1.69} \\
    \rowcolor{blue!15} \cellcolor{white}
    & \cellcolor{white} & \textbf{MUMINS $\Delta\mathbf{b} = \mathbf{0}$} & \textbf{1.13 $\pm$ 0.44} & \textbf{30.87 $\pm$ 2.77} & \textbf{95.34 $\pm$ 2.71} &  4.08 $\pm$ 1.76 & 25.80 $\pm$ 3.12  \\
    \rowcolor{blue!15} \cellcolor{white}
    & \cellcolor{white} & \textit{\textbf{MUMINS} $\Delta\mathbf{b} = \mathrm{GT}$} & \textit{1.05 $\pm$ 0.34} & \textit{31.35 $\pm$ 2.37} & \textit{95.64 $\pm$ 2.28} & \textit{3.69 $\pm$ 1.15} & \textit{26.38 $\pm$ 2.47} \\
\bottomrule
\end{tabular}
\end{table*}

We evaluate MUMINS against GM-AE~\cite{GMAE}, NGP-Net~\cite{NGP}, and TADM-3D~\cite{TADM3D} on PNG~\cite{NGP} lung CT, and BrLP~\cite{BrLP} and TADM-3D on OASIS-3~\cite{OASIS3} brain MRI, plus a no-change copy baseline ($\hat{\mathbf{x}}^{(\mathrm{future})}=\mathbf{x}$, i.e.\ $\hat{\mathbf{y}}_0=\mathbf{0}$).

Table~\ref{tab:results} reports both blur-conditioning settings (Sec.~\ref{sec:conditioning}). MUMINS faithfully follows whatever $\Delta\mathbf{b}$ it is given ($r=0.974$--$0.980$, Supp.~Fig. G.4), including $\Delta\mathbf{b}=\mathbf{0}$, which preserves the baseline's own blur. These retrospective benchmarks, however, contain real inter-visit blur drift that $\Delta\mathbf{b}=\mathbf{0}$ does not chase, so it generally trails $\Delta\mathbf{b}=\mathrm{GT}$ (Supp.~Sec. G). Unless stated otherwise, MUMINS refers to the deployable $\Delta\mathbf{b}=\mathbf{0}$ setting, for parity with baselines that lack access to the true follow-up.

On PNG, MUMINS ($\Delta\mathbf{b} = \mathbf{0}$) attains the best aggregate whole-image MAE, PSNR, and SSIM among all lung methods ($23.53$\,dB PSNR vs.\ $22.88$\,dB for the runner-up GM-AE and $22.78$\,dB for NGP-Net, despite the latter's extra scan), and the best aggregate nodule-ROI PSNR ($24.89$\,dB), with ROI MAE second only to GM-AE ($5.26$ vs.\ $5.17\times10^{-2}$). The \textit{copy} baseline trails all methods but TADM-3D on MAE/PSNR, confirming change is rewarded over copying. Whole-image gains hold across subgroups: MUMINS has the best PSNR and SSIM in every subgroup and the best MAE everywhere except Stable, where GM-AE leads ($3.71$ vs.\ $3.79\times10^{-2}$). ROI-level accuracy is more sensitive: MUMINS clearly leads in Grow ($26.12$\,dB PSNR$_\mathrm{ROI}$ vs.\ $25.18$ for GM-AE), whereas in Stable and Shrink GM-AE's kinetic-growth model attains the best ROI MAE, and the best PSNR$_\mathrm{ROI}$ goes to NGP-Net in Stable ($22.66$ vs.\ $22.43$) and to GM-AE in Shrink ($24.75$ vs.\ $24.26$). Both gaps are consistent with acquisition-blur mismatch: they close on both ROI metrics with $\Delta\mathbf{b}=\mathrm{GT}$ ($24.85$ and $25.45$\,dB PSNR$_\mathrm{ROI}$). With the oracle $\Delta\mathbf{b}=\mathrm{GT}$, the aggregate margin over the best competing method widens to $+2.62$\,dB PSNR and $+2.10$\,dB PSNR$_\mathrm{ROI}$.

On OASIS-3, MUMINS ($\Delta\mathbf{b}=\mathbf{0}$) is essentially matched with the near-saturated \textit{copy} baseline on whole-image metrics since baseline and follow-up differ subtly in most pairs, and ahead of TADM-3D and BrLP on all three (Table~\ref{tab:results}). At the ROI level, TADM-3D holds a small edge: MUMINS leads on MCI ($3.42$ vs.\ $3.69\times10^{-2}$ MAE$_\mathrm{ROI}$) but trails on CN ($4.16$ vs.\ $3.82$), AD ($4.18$ vs.\ $3.79$), and in aggregate ($4.08$ vs.\ $3.80$), with all gaps well within one standard deviation. Supplying $\Delta\mathbf{b}=\mathrm{GT}$ lifts MUMINS's aggregate ROI to its best reported scores, matching or leading TADM-3D on CN, MCI, and aggregate, with TADM-3D remaining stronger on AD ($3.79$ vs.\ $4.27\times10^{-2}$ MAE$_\mathrm{ROI}$), where atrophy produces the largest anatomical change.

Fig.~\ref{fig:NGPReuslts} shows qualitative results of all methods. On lung CT, MUMINS yields the sparsest, lowest-amplitude error maps and shows little error over the nodule; TADM-3D spreads high error beyond the localized errors of GM-AE and NGP-Net. On brain MRI, all methods reconstruct well, so error maps use a much finer scale: MUMINS produces the cleanest maps overall, BrLP yields visibly blurrier follow-ups (likely from its latent-space compression) with error concentrated around the central ventricles and outer cortical boundary, and TADM-3D distributes error more diffusely. Supp.~Sec. I shows failure cases, such as abrupt nodule growth that MUMINS under-predicts.

\begin{table}[t]
\centering
\scriptsize
\setlength{\tabcolsep}{5pt}
\caption{Uncertainty evaluation on both test sets, for the image and the ROI (nodule for CT, brain for MRI). Rows compare single-pass $\boldsymbol{\sigma}$ and $20$-seed MC std, each against voxel-wise error, and their agreement (\textbf{X}). $\rho$: Spearman; $\mathrm{PICP}_{95}\!\to\!0.95$; mean\,$\pm$\,std.}
\label{tab:uncertainty}
\begin{tabular}{@{}ll cc cc@{}}
\toprule
& & \textbf{Lung CT}$_{\text{Img}}$ & \textbf{Lung CT}$_{\text{ROI}}$ & \textbf{Brain MRI}$_{\text{Img}}$ & \textbf{Brain MRI}$_{\text{ROI}}$ \\
\midrule
\multirow{4}{*}{\rotatebox[origin=c]{90}{\textbf{Pred $\boldsymbol{\sigma}$}}}
 & $\rho\uparrow$    & $0.61{\pm}0.15$ & $0.13{\pm}0.28$ & $0.54{\pm}0.05$ & $0.26{\pm}0.07$ \\
 & AUSE$\downarrow$  & $0.02{\pm}0.01$ & $0.06{\pm}0.05$ & $0.002{\pm}0.001$ & $0.03{\pm}0.01$ \\
 & $\mathrm{PICP}_{95}$ & $0.91{\pm}0.06$ & $0.80{\pm}0.22$ & $0.98{\pm}0.02$ & $0.97{\pm}0.04$ \\
 & MPIW              & $0.45{\pm}0.01$ & $0.46{\pm}0.02$ & $0.44{\pm}0.00$ & $0.47{\pm}0.01$ \\
\midrule
\multirow{4}{*}{\rotatebox[origin=c]{90}{\textbf{MC std}}}
 & $\rho\uparrow$    & $0.70{\pm}0.14$ & $0.17{\pm}0.29$ & $0.56{\pm}0.06$ & $0.32{\pm}0.08$ \\
 & AUSE$\downarrow$  & $0.01{\pm}0.01$ & $0.05{\pm}0.05$ & $0.001{\pm}0.001$ & $0.02{\pm}0.01$ \\
 & $\mathrm{PICP}_{95}$ & $0.78{\pm}0.14$ & $0.61{\pm}0.23$ & $0.20{\pm}0.05$ & $0.59{\pm}0.14$ \\
 & MPIW              & $0.22{\pm}0.08$ & $0.30{\pm}0.20$ & $0.04{\pm}0.00$ & $0.12{\pm}0.01$ \\
\midrule
\multirow{1}{*}{\rotatebox[origin=c]{90}{\textbf{X}}}
 & $\rho\uparrow$    & $0.79{\pm}0.08$ & $0.50{\pm}0.26$ & $0.73{\pm}0.02$ & $0.65{\pm}0.03$ \\
\bottomrule
\end{tabular}
\end{table}

\noindent\textbf{Uncertainty.} Table~\ref{tab:uncertainty} scores the single-pass predicted $\boldsymbol{\sigma}$ against voxel-wise error, over the image and the ROI (nodule/brain). The meaningful measure differs by dataset. For lung, the $64^3$ volumes are cropped around the nodule, so whole-image scores are anatomically meaningful: $\boldsymbol{\sigma}$ localizes error well ($\rho=0.61$) and is reasonably calibrated ($\mathrm{PICP}_{95}\,0.91$), far cheaper than the $20\times$-costlier MC reference ($\rho=0.70$, $\mathrm{PICP}_{95}\,0.78$). The nodule ROI is a small sub-region, so its correlations are unstable for both $\boldsymbol{\sigma}$  ($\rho=0.13{\pm}0.28$) and MC ($\rho=0.17{\pm}0.29$) and not meaningfully interpretable. For the brain, volumes are skull-stripped, inflating whole-image scores with the large zero background, so uncertainty inside the ROI is the meaningful measure. There, reconstruction is near-saturated, and error is small and diffuse, leaving little structure to localize: MC's repeated sampling gives a modest localization edge ($\rho=0.32$ vs.\ our $0.26$), but calibration is the discriminating axis, where the single diffusion process uncertainty $\boldsymbol{\sigma}$ remains better calibrated ($\mathrm{PICP}_{95}\,0.97$) while MC severely under-covers ($0.59$). Across both datasets, both estimates are correlated (cross-$\rho$ $0.79$ lung, $0.65$ brain ROI), so a single pass recovers much of the uncertainty structure MC obtains only by repeated sampling. This structure could stem from the learned variance head or the sampler's injected stochasticity; setting $\eta=0$ removes the latter, yet $\boldsymbol{\sigma}$ remains error-correlated, confirming the variance head learns meaningful uncertainty on its own (Supp.~Sec. F).

\begin{figure}[htbp]
    \centering
    \includegraphics[width=0.41\textwidth]{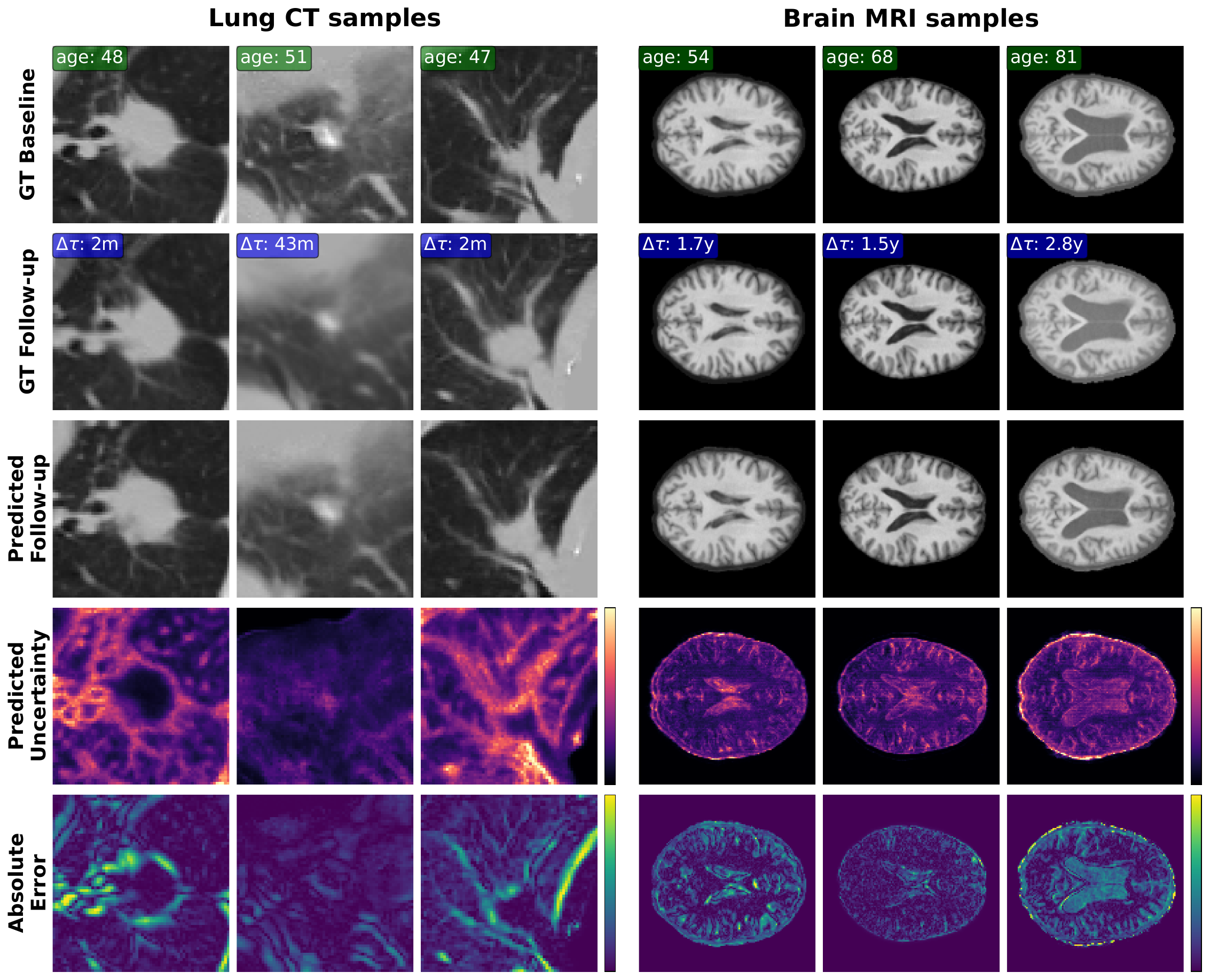} 
    \caption{MUMINS follow-up and uncertainty predictions (test samples). $\Delta\tau$ denotes inter-scan interval (months \texttt{m}, years \texttt{y}).}
    \label{fig:uncertainty}
\end{figure}

\begin{figure}[htbp]
    \centering
    \begin{overpic}[width=0.34\textwidth]{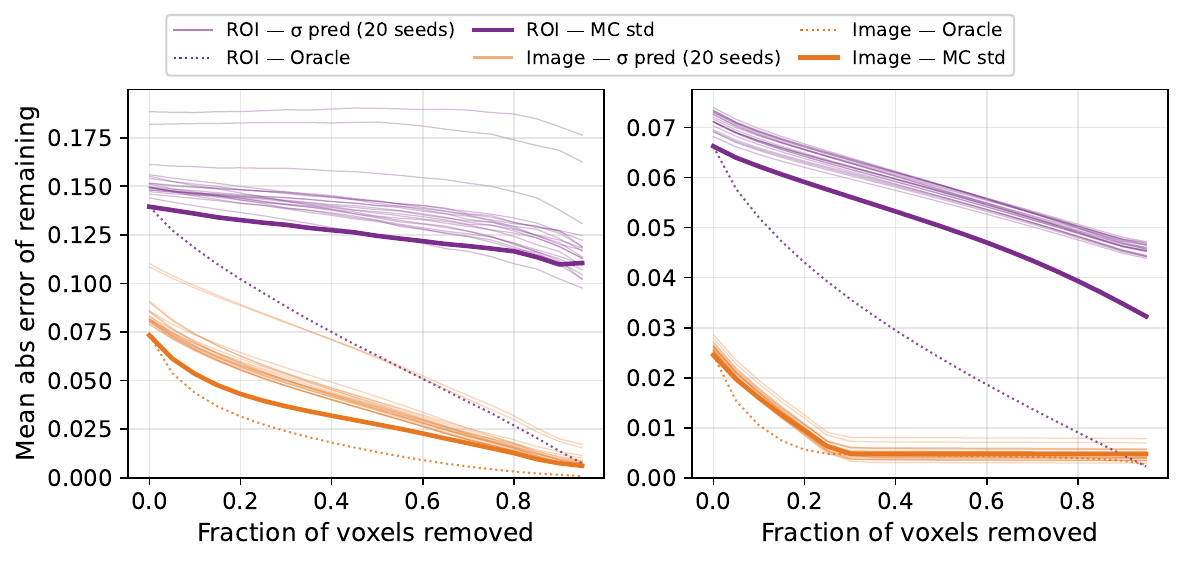}
        \put(28, -4){\textbf{\footnotesize (a)}}  
        \put(75, -4){\textbf{\footnotesize (b)}} 
    \end{overpic}
    \caption{Sparsification curves on PNG lung CT (a) and OASIS brain MRI (b) test sets; ROI is the nodule (a) and brain (b). }
    \label{fig:sparsification}
\end{figure}

Figure~\ref{fig:uncertainty} shows predicted uncertainty and absolute-error maps for representative cases (axial, sagittal, and coronal views in Supp. Fig. H.6). Sparsification curves (Fig.~\ref{fig:sparsification}) show both estimators' MAE dropping monotonically toward the oracle as high-uncertainty voxels are removed first; the single-pass curve sits above the 20-seed MC reference, consistent with its higher AUSE (Table~\ref{tab:uncertainty}).

\newcommand{\cmark}{\ding{51}}
\newcommand{\xmark}{\ding{55}}

\subsection{Ablation Studies}
\label{sec:ablations}
We ablate MUMINS on the PNG lung validation split~\cite{NGP}; the $8\times$-larger brain ($128^3$) configuration precludes exhaustive sweeps. Table~\ref{tab:ablation_v2} isolates five components on a shared 3D U-Net backbone~\cite{cciccek20163d} and training iterations: joint diffusion (JD), residual prediction (RP), blur conditioning (BC), uncertainty prediction (UP), and baseline re-injection (BR), starting from a single-channel diffusion in which the baseline enters only as conditioning (JD-free), as in typical conditional diffusion models for progression~\cite{BrLP, TADM3D, TaDiff}. Predicting the image directly (rows~1--2) yields high variance, although JD already improves on the baseline in every metric; RP sharply tightens this spread (PSNR std $4.09\!\to\!1.65$\,dB), and BC then lifts fidelity (PSNR $17.40\!\to\!20.69$\,dB), with UP adding a further, essentially cost-free gain ($21.00$ vs.\ $20.69$\,dB). BR, a parameter-free inference-time mechanism, adds the final large step (MAE $5.55\!\to\!3.86\times10^{-2}$, PSNR $21.00\!\to\!23.73$\,dB, SSIM $56.27\%\!\to\!72.38\%$), the largest SSIM gain of any component. Full analysis, qualitative examples, and sampler sweeps ($\zeta$, $\hat{s}$, $\eta$) are in Supp.~Sec. F.
\begin{table}[htbp]
\centering
\footnotesize
\setlength{\tabcolsep}{2pt}
\caption{Ablation study of MUMINS variants on the PNG validation set~\cite{NGP}. Starting from an image-only conditional baseline (top row), we add joint diffusion (JD), residual prediction (RP), acquisition-blur difference conditioning (BC) with $\Delta\mathbf{b} = \mathbf{0}$ inference, uncertainty prediction (UP) and baseline reinjection (BR).}
\label{tab:ablation_v2}
\begin{tabular}{ccccc ccc}
\toprule
\textbf{JD} & \textbf{RP} & \textbf{BC} & \textbf{UP} & \textbf{BR} & \textbf{MAE ($\times 10^{-2}$)} $\downarrow$ & \textbf{PSNR (dB)} $\uparrow$ & \textbf{SSIM ($\%$)} $\uparrow$ \\
\midrule
\xmark & \xmark & \xmark & \xmark & \xmark & 19.76 $\pm$ 10.82 & 14.19 $\pm$ 5.43  & 41.84 $\pm$ 26.55 \\
\cmark & \xmark & \xmark & \xmark & \xmark & 11.54 $\pm$ 5.33  & 16.98 $\pm$ 4.09  & 54.28 $\pm$ 19.84 \\
\cmark & \cmark & \xmark & \xmark & \xmark & 8.62  $\pm$ 1.68  & 17.40 $\pm$ 1.65  & 45.45 $\pm$ 10.15 \\
\cmark & \cmark & \cmark & \xmark & \xmark & 5.75  $\pm$ 1.60  & 20.69 $\pm$ 2.29  & 54.64 $\pm$ 11.42 \\
\cmark & \cmark & \cmark & \cmark & \xmark & 5.55  $\pm$ 1.70  & 21.00 $\pm$ 2.59  & 56.27 $\pm$ 12.49 \\
\rowcolor{blue!15}
\cmark & \cmark & \cmark & \cmark & \cmark & \textbf{3.86 $\pm$ 1.69} & \textbf{23.73 $\pm$ 3.98} & \textbf{72.38 $\pm$ 13.65} \\
\bottomrule
\end{tabular}
\end{table}

\section{Conclusions}
We presented MUMINS, a diffusion framework that synthesizes a follow-up scan and a voxel-wise uncertainty map in a single reverse diffusion process. Modeling a joint baseline-residual state with per-step baseline re-injection preserves fine anatomy while generating plausible change; a likelihood-based variance head yields uncertainty without Monte Carlo's costlier repeated sampling. With no organ-specific components, MUMINS matches or outperforms domain-specific baselines on lung CT and brain MRI, with single-pass uncertainty correlating cheaply with the MC reference. Two limitations remain: single-pass localization trails the costlier MC
reference, and predictions queried at different $\Delta\tau$ from the same baseline are generated independently, with no explicit mechanism enforcing cross-horizon consistency. Efficient computation of the denoiser's volumetric Jacobian for uncertainty estimation is a promising direction, as is extending MUMINS to further organs and modalities toward a general progression-synthesis model.
 \section*{Acknowledgments}
A.~Oliveras acknowledges Eurecat's ``Vicente L\'opez'' PhD grant program.
This work was supported by computational resources provided by Finis Terrae III
(CESGA) and the Barcelona Supercomputing Center (BCV-2026-2-0027), and by the
PHASE IV AI project, funded by the European Union under the Horizon Europe grant
agreement No.~101095384.
This work was partially funded by Icrea Academia'2022 (Generalitat de Catalunya),
IDEATE (PID2022-141566NB-I00) and TRI-VISION (PID2025-173459NB-C21) both funded by
MICIU\slash AEI\slash 10.13039\slash 501100011033 and by FEDER\slash EU, and
EXPLORA-SCALE (AIA2025-163919-C51) funded by
MICIU\slash AEI\slash 10.13039\slash 501100011033.
B.~Nagarajan acknowledges AI4S fellowship within the ``Generaci\'on D'' initiative
by Red.es, Ministerio para la Transformaci\'on Digital y de la Funci\'on P\'ublica,
for talent attraction (C005/24-ED CV1), funded by NextGenerationEU through PRTR.

{
    \small
    \bibliographystyle{ieeenat_fullname}
    \bibliography{main}
}

\end{document}